\documentclass{article}
\usepackage{spconf,amsmath,graphicx}
\usepackage{multirow}
\usepackage{amssymb}

\def\L{{\cal L}}

\title{Frame Stacking and Retaining for Recurrent Neural Network Acoustic Model}
\name{Xu Tian, Jun Zhang, Zejun Ma, Yi He, Juan Wei}

\address{Alibaba Shenma Search, Beijing, China\\
	\small \texttt\{xu.tian, zj102217, zejun.mamzj, heyi.hy, wj80290\}@alibaba-inc.com}

\begin{document}

\maketitle

\begin{abstract}
Frame stacking is broadly applied in end-to-end neural network training like connectionist temporal classification (CTC), and it leads to more accurate models and faster decoding. However, it is not well-suited to conventional neural network based on context-dependent state acoustic model, if the decoder is unchanged. In this paper, we propose a novel frame retaining method which is applied in decoding. The system which combined frame retaining with frame stacking could reduces the time consumption of both training and decoding. Long short-term memory (LSTM) recurrent neural networks (RNNs) using it achieve almost linear training speedup and reduces relative 41\% real time factor (RTF). At the same time, recognition performance is no degradation or improves sightly on Shenma voice search dataset in Mandarin.

\end{abstract}

\section{Introduction}
\label{sec:intro}

In the last few years, deep neural networks (DNNs) combined with hidden Markov models (HMMs) have been widely employed in acoustic modeling for large vocabulary speech recognition \cite{hinton2012deep}. More recently, Recurrent neural networks (RNNs), especially long short-term memory (LSTM) RNNs, have been shown to outperform DNNs \cite{graves2013speech,graves2013hybrid,sak2014long}. 

DNNs always stack the fixed number of neighboring frames feature together as a new feature of current frame, as lack of temporal information. Frame stacking is a effective way that DNNs could learn past and future context knowledge \cite{grezl2008optimizing,thomas2009phoneme,vanhoucke2013multiframe}. Though RNNs are able to remember long-term information, frame stacking could also provide useful contextual information \cite{wollmer2011feature}. For small LSTM network, it employs the same stacking method with DNNs, and the number of frames is no reduction. With the growth of LSTM models size, the influence of contextual information provided by frame stacking fades gradually.

However, the neural networks combined with connectionist temporal classification (CTC) criterion gives the frame stacking rebirth \cite{sak2015fast,sak2015acoustic}. Connectionist temporal classification (CTC) criterion provides a mechanism to learn an neural network while mapping a input frame sequence to a output label sequence \cite{graves2006connectionist}. The length of output sequence could be much more shorter than that of input sequence, because of the blank symbol of CTC. Thus, there is no need of a frame-level alignment for cross-entropy (CE) training. CTC-LSTM acoustic models using context dependent phones (CD-phones) perform as well as conventional models \cite{senior2015context}. As it utilizes a larger modeling unit, the successive frames could be stacked together as a super frame. If we regard DNNs frame stacking with a sliding window method, its sliding step is one. The frame stacking of CTC-LSTM is more flexible that its sliding step could be longer. Even the sliding step could be equal to window length, and there is no overlap between two windows. As a result, the frame stacking reduce the frame rate, and leads to faster training and decoding.

The traditional RNN models, which are still competitive, could also utilize the frame stacking directly in the training phase. But it brings prominent deterioration of decoding result, if the decoding network is unchanged. It is an intuitional way to remodel HMM structure in order to match the modeling unit, and decoding network is needed to rebuild correspondingly \cite{pundak2016lower}. In this paper, we explore conventional RNN models using frame stacking, and propose a novel frame retaining method which is applied in decoding phase and keeps the original decoding network. Frame stacking and retaining will be describe in Section~\ref{sec:fs}. LSTM models are successfully trained on large scale dataset in Section~\ref{sec:exp}, followed by conclusions in Section~\ref{sec:conclusion}.

\section{Frame Stacking and Retaining}
\label{sec:fs}

\subsection{Non-overlapping Frame Stacking}

\begin{figure}[t]
	\label{fig:fs}
	\centering
	\includegraphics[width=3in]{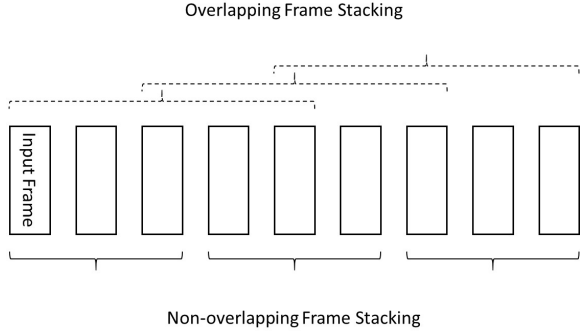}
%	\\[-.8in]
	\caption{\label{fig:stacking}Overlapping and non-overlapping frame stacking.}
%	\vspace{0.8in}
\end{figure}

In the conventional acoustic modeling systems, features is extracted with frame segmentation, and they are computed every fixed steps on fixed frame windows. Frame stacking is a kind of frame re-segmentation, which stacks temporal neighboring frames to a super frame. There is two kinds of frame stacking, overlapping one and non-overlapping one, as shown in Figure~\ref{fig:stacking}. They could brings linear reduction of input frames, and the degree of it depends on the shift step of overlapping one or the frame window of non-overlapping one. Since the original feature is extracted with sliding frame window, there is no need to use sliding window in frame stacking again. Therefore, we prefer non-overlapping frame stacking for RNNs, which has temporal memory structure.

For speech recognition applications, DNNs input frames always contain context information through packing temporal sequential left and right frames together. It could cover the shortage of no dynamic temporal behavior.

In contrast, RNNs do not need to pack the context information to obtain the sequential ability. It stacks neighboring frames to a super frame, because of the stationarity of the speech signal. Super frames provides multiple frames information as the new input of the network with no information missing, so the quantity of input frames decreases linearly. The super frame's label comes from the label of middle frame of successive frames. The network only needs to enlarge this architecture properly, and the main enlargement is for the input layer. As a result, the time cost of network training decreases almost linearly.

\subsection{Frame Retaining in Decoding}
\begin{figure}[t]
	\label{fig:d}
	\centering
	\includegraphics[width=3in]{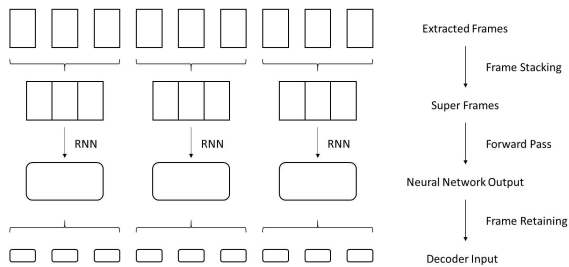}
%	\\[-.8in]
	\caption{\label{fig:decoding}Decoding with frame retaining.}
%	\vspace{0.8in}
\end{figure}

Frame stacking could substantially reduce the training time, and it could also have the same effect on decoding phase. It has been demonstrated in CTC systems \cite{sak2015fast}. As CTC is phone-level modeling, the granularity after stacking is still suitable for CTC decoding. But conventional RNNs is state-level modeling and weighted finite state transducer (WFST) is state-level correspondingly, so the granularity is too large to decode. In order to maintain the decoding granularity, frame retaining is proposed as presented in Figure~\ref{fig:decoding}.

The size of frame stacking window is denoted as $N$. After $N$ successive frames are extracted in a signal stream, they are stacked to a super frame in the same way of training phase. Consequently, the super frame retains for $N$ frames time with frame retaining method. The neighboring frame has similar properties, so super frame represents them, after aggregating their features. In traditional decoding method, features of each frame needs to pass through the network, and $N$ frames mean $N$ times forward passes. But a super frame passes through the network only once carrying the all information of $N$ frames, and the result of the super frame's forward pass is multiplexed at the rest of frame time. Moreover, WFST does not need to rebuild for frame stacking. Thus, decoder spends less time in general, as $N-1$ times of forward pass is skipped and computation consumption of one forward pass increases only a little. 

\subsection{Acoustic Model Trained with Cross-Entropy}
Let $\mathbf{x}=x_1,\dots,x_T$ denote a input sequence of $T$ acoustic feature vectors, where $x_t\in \mathbb{R}^N$, and $\mathbf{w}$ an output word sequence. The acoustic likelihood is decomposed as follows:
$$p(\mathbf{x|w})=\prod_{t=1}^{T}p(x_t|l_t)p(l_t|l_{t-1})$$
where $l_1,\dots,l_T$ is the label sequence, which is obtained by existing models. In the hybrid decoding, the emission probability of HMM is represented as $p(x_t|l_t)=p(l_t|x_t)p(x_t)/p(l_t)$. The label posterior is given by the output of a neural network acoustic model, and it could be computed using a context of $N$ frames with frame stacking. The label prior $p(l_t)$ is counted by the label of existing model's alignment.

The acoustic model of neural network is first trained to maximize the cross-entropy (CE) loss with the input sequence $\mathbf{x}$ and the corresponding frame-level alignment $\mathbf{l}$, as follow:
$$\L_{CE}=-\sum_{(\mathbf{x},\mathbf{l})}\sum_{t=1}^{T}\log p(l_t|x_t)$$
Where $p(l|x)$ is the label posterior after the softmax output layer of the neural network.

\subsection{Sequence Discriminative Training}
CE provides a kind of frame-wise discriminative training criterion, but it not enough for speech recognition which is a sequence problem. Sequence discriminative training using state-level minimum bayes risk (sMBR) has shown to further improve performance of neural networks first trained with CE \cite{kingsbury2009lattice,sak2015learning}. The model first trained by CE loss is frame-level accurate, and it is further trained with sMBR to get sequence-level accuracy. Frame stacking and retaining are also applied in sMBR training. It also gets almost linear speedup. Moreover, on the basis of frame-level accurate model, only a part of dataset is needed for sMBR training.

\section{Experiments and Results}
\label{sec:exp}

\subsection{Experiments Setup}

The neural network is trained on 17000 hours dataset which is collected from Shenma voice search. It is one of the most popular mobile search engines in China. The dataset is created from anonymous online users' search queries in Mandarin, and all audio file's sampling rate is 16kHz, recorded by mobile phones. This dataset consists of many different conditions, such as diverse noise even low signal-to-noise, babble, dialects, accents, hesitation and so on. 

The dataset is divided into training set, validation set and test set separately, and the quantity of them is shown in Table~\ref{table:data}. The three sets are split according to speakers, in order to avoid utterances of same speaker appearing in three sets simultaneously. The test sets of Shenma voice search are called Shenma Test.

\begin{table}[t]
	
	\centering	
	\begin{tabular}{cc}
		\hline
		\textbf{Dataset} & \textbf{Hours}   \\ 
		\hline
		Training set  & 16150  \\
		Validation set & 850   \\
		Test set &  10  \\
		Total & 17010  \\
		\hline
	\end{tabular}	
    \caption{ \label{table:data} \small The time summation of different sets of Shenma voice search.}
\end{table}

LSTM RNNs outperform conventional RNNs for speech recognition system, especially deep LSTM RNNs, because of its long-range dependencies more accurately for temporal sequence conditions \cite{hermans2013training,sak2015acoustic}. Shenma voice search is a streaming service that intermediate recognition results displayed while users are still speaking. So as for online recognition in real time, we prefer unidirectional LSTM model rather than bidirectional one. Thus, the training system is unidirectional LSTM-based.

A 26-dimensional filter bank and 2-dimensional pitch feature is extracted for each frame, and is concatenated with first and second order difference as the final input of the network. The extraction happens every 10 ms with 25 ms frame window. The architecture we trained consists of two LSTM layers with sigmoid activation function, followed by a full-connection layer. The out layer is a softmax layer with 11088 hidden markov model (HMM) tied-states as output classes, the loss function is CE. After CE training, the model is trained with sMBR. The performance metric of the system in Mandarin is reported with character error rate (CER). The alignment of frame-level ground truth is obtained by GMM-HMM system. Mini-batched SGD is utilized with momentum trick and the network is trained for a total of 4 epochs. 
 5-gram language model is leveraged in decoder, and the vocabulary size is as large as 760000.

It has shown that blockwise model-update filtering (BMUF) outperforms traditional model averaging method, and it is utilized at the synchronization phase \cite{chen2016scalable}. Its block learning rate and block momentum are set as 1 and 0.9. After synchronizing with BMUF, exponential moving average (EMA) method further updates the model in non-interference way \cite{tian2017}. The training system is deployed on the MPI-based HPC cluster where 8 GPUs. Each GPU processes non-overlap subset split from the entire large scale dataset in parallel. 

Local models from distributed workers synchronize with each other in decentralized way. In the traditional model averaging and BMUF method, a parameter server waits for all workers to send their local models, aggregate them, and send the updated model to all workers. Computing resource of workers is wasted until aggregation of the parameter server done. Decentralized method makes full use of computing resource, and we employ the MPI-based mesh AllReduce method \cite{tian2017}. It is significant to promote training efficiency, when the size of neural network model is too large. The EMA model is also updated additionally, but not broadcasting it.

%\begin{figure}[t]
%	%	\label{fig:fs}
%	\centering
%	\includegraphics[width=3in]{fig/mesh_allreduce.png}
%	%	\\[-.8in]
%	\caption{\label{fig:allreduce}Mesh AllReduce of model averaging.}
%	%	\vspace{0.8in}
%\end{figure}

\subsection{Results}

Frame stacking cuts down the number of input frames, so it leads to almost linear speedup of training. But when its model is applied in the decoder directly, it will cause the great CER degradation, as the modeling duration does not match. The decoding network is generated to fit for original modeling duration. $N$ frames corresponds only one input feature vectors of decoder for frame stacking, while $N$ frames corresponds $N$ of them for original modeling. Therefore, frame retaining in decoder could match the number of input feature vectors and that of frames. We denote the number of non-overlapping stacked frames as $FS$, and the times of a super frame retaining as $FR$. As shown in Table~\ref{table:retain}, if a super frame is stacked by 3 frames for 4-layers LSTM models, $FR=1$ increase relative 415\% CER, and the other modeling duration mismatch also results in worse performance of decoder. $FR$ being 1 means no frame retaining, and it demonstrates that only frame stacking could not improve the accuracy of non-CTC neural network.

\begin{table}[t]
	\centering	
	\begin{tabular}{ccc}
		\hline
		\textbf{$FS$} & \textbf{$FR$} & \textbf{CER}   \\ 
		\hline
		3  & 1 & 19.64  \\
		3 &  2 & 4.53  \\
		3 & 3 & 3.81 \\
		\hline
	\end{tabular}	
	\caption{ \label{table:retain} \small the CERs of different FR with the same FS for a 4-layers LSTM model.}
\end{table}

Frame stacking and retaining not only spends less time in training, but also brings faster decoder. Real time factor is utilized to evaluate the decoding speed. CERs and RTFs of 4-layers LSTM models with different number of stacked frames and matching frame retaining are presents in Table~\ref{table:stack}. Neighboring frames have similar features, so there is no information omitted in stacking process. It does not reduce the performance of recognition, and even improves it, as shown in Table~\ref{table:stack}. For our system, it is optimal that $FS$ and $FR$ are both set as 3. It reduces relative 41\% RTF, and accuracy improves sightly.

\begin{table}[t]
	\centering	
	\begin{tabular}{cccc}
		\hline
		\textbf{$FS$} & \textbf{$FR$} & \textbf{CER} & \textbf{RTF}   \\ 
		\hline
		1  & 1 & 3.89 & 0.41 \\
		2 & 2 & 3.85 & 0.29 \\
		3 &  3 & 3.81 & 0.24 \\
		4 & 4 & 4.23 & 0.25 \\
		\hline
	\end{tabular}	
	\caption{ \label{table:stack} \small 4-layers LSTM models with different number of stacked frames and matching frame retaining.}
\end{table}

\section{Conclusion}
\label{sec:conclusion}
In this work, we propose frame retaining in conventional neural networks with frame stacking. The parameters of frame stacking and retaining should be equal, in order that they have the same modeling duration. It leads to almost linear training speedup and faster decoding, while the performance of speech recognition does not decrease. Unidirectional LSTM models are trained to verify it on large scale speech recognition. RTF reduces relative 41\% and the character accuracy improves sightly compared with no use of frame stacking and retaining.

%\vfill\pagebreak

\bibliographystyle{IEEEbib}
\bibliography{ref}

\end{document}